\documentclass[twoside,11pt]{article}

\usepackage{jmlr2e}
\usepackage{amsfonts}
\usepackage{amssymb}
\usepackage{amstext}
\usepackage{amsmath,bm}
\usepackage{natbib}
\usepackage{graphicx}
\usepackage{booktabs}
\usepackage{array}
\usepackage{multirow}
\usepackage{algorithm}
\usepackage{algorithmic}
\usepackage{mathrsfs}
\usepackage{subfigure}
\usepackage{epsfig}
\usepackage{multicol}
\usepackage{makecell}

\jmlrheading{13}{2012}{1059-1062}{8/11}{4/12}{Zhao, Liu, Roeder, Lafferty and Wasserman}

\ShortHeadings{High-dimensional Undirected Graph Estimation}{Zhao, Liu, Roeder, Lafferty and Wasserman}
\firstpageno{1}

\begin{document}

\title{The \texttt{huge} Package for High-dimensional Undirected Graph Estimation in \texttt{R}}

\author{\name Tuo Zhao \email tourzhao@jhu.edu\\
       \name Han Liu \email hanliu@cs.jhu.edu \\
       \addr Johns Hopkins University\\
       Baltimore, MD 21218, USA
       \AND
       \name Kathryn Roeder \email roeder@stat.cmu.edu\\
       \name John Lafferty \email lafferty@cs.cmu.edu\\
       \name Larry Wasserman \email larry@stat.cmu.edu\\
       \addr Carnegie Mellon University\\
       Pittsburgh,PA, 15213}

\editor{Mikio Braun}

\maketitle

\begin{abstract}
We describe an \texttt{R} package called \texttt{huge} (ver 1.1.2),
that provides easy-to-use functions for estimating high dimensional
undirected graphs from data.  This package implements recent results
in the literature, including 
\citet{Friedman07a}, \citet{Liu09} and \citet{Liu10}.  Compared with \texttt{glasso}, the \texttt{huge} package provides
several extra features: (i) instead of using \texttt{Fortran}, it is
written in \texttt{C}, which makes the code more portable and easier
to modify; (ii) besides fitting Gaussian graphical models, it also
provides functions for fitting high dimensional  semiparametric
Gaussian copula models, data-dependent model selection, data
generation and graph visualization; and  (iii) to achieve better
scalability, it incorporates correlation screening into graph
estimation. In particular, the package allows the user to apply both
lossless and lossy screening rules to scale up for high-dimensional problems,
making a tradeoff between computational and statistical efficiency.    
\end{abstract}

\vspace{0.1in}
\begin{keywords}
high-dimensional undirected graph estimation, glasso,  huge, semiparametric graph estimation, data-dependent model selection, lossless screening, lossy screening. 
\end{keywords}

\section{Overview}

Significant progress has been made recently on designing efficient
algorithms to learn undirected graphical models from high-dimensional
observational datasets. Existing packages include \texttt{glasso}
\citep{Friedman07a}, \texttt{Covpath} and
\texttt{CLIME}.  In particular, the \texttt{glasso}
package has been widely adopted by statisticians and computer
scientists due to its friendly user-inference and efficiency.  In this
paper, we describe a newly developed \texttt{R} package named
\texttt{huge}\footnote{The development was partially supported by google summer of code project} (High-dimensional Undirected Graph Estimation).
Compared with \texttt{glasso}, the core engine of \texttt{huge} is
coded in \texttt{C}, making modifications of the package more accessible to researchers
from the computer science and signal processing communities. The
package includes a wide range of functional modules, including data
generation, data preprocessing, graph estimation, model selection, and
visualization. Many recent methods have been implemented, including
the nonparanormal \citep{Liu09} for estimating a high dimensional
Gaussian copula graph, the StARS \citep{Liu10} approach for
stability-based graphical model selection, and correlation screening for high dimensional graph estimation. The package supports two modes of
screening, lossless \citep{Witten11} and lossy screening.  When using lossy screening, the user
can select the desired screening level to scale up for high-dimensional problems, but this introduces some estimation bias.

\section{Design and Implementation}

\begin{figure}[!htb]
\centering
\includegraphics[width=5.5in]{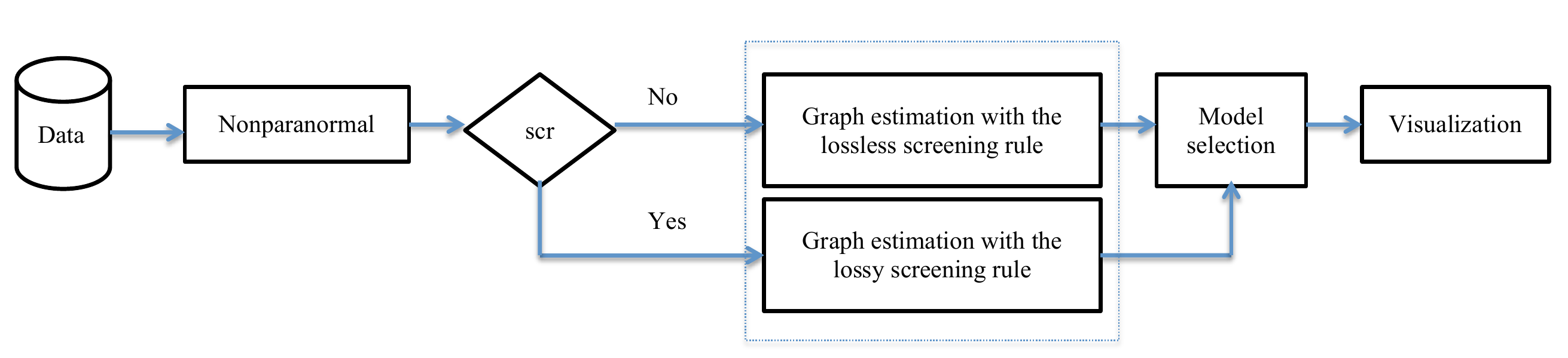}
\vspace{-0.2in}
\caption{\small The graph estimation pipeline.}
\label{flowchart}
\end{figure}

The package \texttt{huge} aims to provide a general framework for high-dimensional undirected graph estimation. The package includes six functional modules (M1-M6), see (Figure \ref{flowchart}).
\vspace{-0.3in}

{\bf M1. Data Generator}: The function \texttt{huge.generator()} can simulate multivariate Gaussian data with different undirected graph structures, including hub, cluster, band, scale-free, and Erd{\"{o}s-R{\'{e}}nyi random graphs. The sparsity level of the obtained graph and signal-to-noise ratio can also be set up by users. 

{\bf M2. Semiparametric Transformation}: The function \texttt{huge.npn()} implements the nonparanormal method \citep{Liu09} for estimating a semiparametric Gaussian copula model. Motivated by  additive models, the nonparanormal family extends the Gaussian distribution by marginally transforming the variables using smooth functions. Computationally, the  nonparanormal transformation only requires one pass through the data matrix.

{\bf M3. Graph Screening}: The \texttt{scr} argument in the main
function \texttt{huge()} controls the use of large-scale correlation
screening before graph estimation. The function supports two types of
screening rules, lossless screening and lossy screening. The lossless
screening method is from \citet{Witten11}.  Such screening procedures can greatly
reduce the computational cost and achieve equal or even better
estimation by reducing the variance at the expense of increased bias.

{\bf M4. Graph Estimation}: Similar to the \texttt{glasso} package,
the \texttt{method} argument in the  \texttt{huge()} function supports
two estimation methods: (i) the Meinshausen-B{\"{u}}hlmann  covariance
selection algorithm \citep{Meinshausen06} and (ii) the graphical lasso
algorithm \citep{Friedman07a}. In our implementation, we
exploit many suggested tricks and practices from \citet{Friedman10a}. For example, we solve each individual
lasso problem using coordinate descent combined with active set and
covariance update tricks. One difference between \texttt{huge} and
\texttt{glasso} is that we implement all the core components using
\texttt{C} instead of \texttt{Fortran}.   The code is also
memory-optimized  using the sparse matrix data structure so that it can
handle larger datasets  when estimating and storing full
regularization paths.  We also provide an additional graph estimation 
method based on thresholding the sample correlation matrix.  Such an approach 
is computationally efficient and has been widely applied in biomedical research. 

{\bf M5. Model Selection}:  The function \texttt{huge.select()}
provides three regularization parameter selection methods: the
stability approach for regularization selection (StARS)
\citep{Liu10}; a modified rotation information criterion (RIC); and the extended Bayesian information criterion. The latter approach is a likelihood-based model
selection criterion that is only applicable for the graphical lasso
method. StARS conducts many subsampling steps to calculate
U-statistics, which is computationally intensive but can be trivially
parallelized.  RIC is closely related to the permutation approach for model selection and scales to large datasets. 

{\bf M6. Graph Visualization}:  The plotting functions
\texttt{huge.plot()} and \texttt{plot()} provide visualizations of the
simulated data sets, estimated graphs and paths. The implementation is
based on the \texttt{igraph} package.  Due to the limits of
\texttt{igraph}, sparse graphs with only up to 2,000 nodes can
be visualized.

\section{User Interface by Example}

We illustrate the user interface by analyzing a stock
market data which we contribute to the \texttt{huge} package\footnote{We thank Mladen Kolar for providing the python code to craw the data from the web.}. We
acquired closing prices from all stocks in the S\&P 500 for all the
days that the market was open between Jan 1, 2003 and Jan 1,
2008. This gave us 1258 samples for the 452 stocks that remained in the S\&P 500 during the entire time period.
\begin{verbatim}
> library(huge)                                           
> data(stockdata)                                           # Load the data
> x = log(stockdata$data[2:1258,]/stockdata$data[1:1257,])  # Preprocessing
> x.npn = huge.npn(x, npn.func="truncation")                # Nonparanormal
> out.npn = huge(x.npn,method = "glasso", nlambda=40,lambda.min.ratio = 0.4)  
\end{verbatim}
Here the data have been transformed by calculating the log-ratio of the
price at time $t$ to price at time $t-1$.
The nonparanormal transformation is applied to the data, and a
graph is estimated using the graphical lasso (the default is the Meinshausen-B\"{u}hlmann estimator). The program automatically sets up a sequence of 40 regularization parameters and estimates the  graph path. The lossless screening method is applied by default. 


\section{Performance Benchmark}
We adopt similar experimental settings as in \citet{Friedman10b} to compare \texttt{huge} with \texttt{glasso} (ver 1.4).  We consider four scenarios with varying sample sizes $n$ and dimensionality $d$, as shown in Table 1. We simulate the data from a normal distribution $N(0,I_d)$. Timings (in seconds) are computed over 10 values of the corresponding regularization parameter, and the range of regularization parameters is chosen so that each method produced approximately the same number of non-zero estimates (the sparsity level is from 0 to about 0.03). The convergence threshold of  both \texttt{glasso} and \texttt{huge} is chosen to be $10^{-4}$. All experiments were carried out on a PC with Intel Core i5 3.3Hz processor. We also tried \texttt{CLIME} (ver 1.0) and \texttt{Covpath} (ver 0.2), but were unable to obtain timing results due to numerical issues.
\begin{table}[!htb]
\footnotesize
  \centering
  \caption{Experimental Results}
  \vspace{0.05in}
  \begin{tabular}[htb!]{l | l | c | r | r | r | r}
   \Xhline{1pt}
    \multicolumn{3}{c|}{\multirow{2}{*}{Method}} & $d=1000$ &$d=2000$ &$d=3000$ &$d=4000$\\[-2pt]
    \multicolumn{3}{l|}{} &$n=100$ & $n=150$ &$n=200$ &$n=300$\\
    \Xhline{1 pt}
    \multicolumn{3}{l|}{\texttt{huge}-Meinshausen-B{\"{u}}hlmann (lossy)} &0.938(0.054) &3.562(0.581) &8.238(0.822) &19.06(1.845)\\
    \multicolumn{3}{l|}{\texttt{huge}-Meinshausen-B{\"{u}}hlmann} &1.247(0.060) &11.88(2.136) &38.86(3.882) &104.0(5.574)\\
    \multicolumn{3}{l|}{\texttt{glasso}-Meinshausen-B{\"{u}}hlmann} &27.91(0.286) &216.9(2.557) &717.0(2.870) &1688(9.991)\\
     \multicolumn{3}{l|}{\texttt{huge}-graphical lasso (lossy)} &23.03(2.261) &222.4(23.92) &709.3(37.68) &1552(86.14)\\
    \multicolumn{3}{l|}{\texttt{huge}-graphical lasso (lossless)} &24.24(2.951) &267.5(39.48) &819.5(47.39) &1750(55.78)\\
    \multicolumn{3}{l|}{\texttt{glasso}-graphical lasso} &79.68(2.338) &742.5(9.831) &2348(12.08) &5455(17.25)\\
    \Xhline{1pt}
  \end{tabular}
\end{table}

For Meinshausen-B{\"{u}}hlmann graph estimation, we can see that
\texttt{huge} achieves the best performance. In particular, when the
lossy screening rule is applied,  \texttt{huge} automatically reduces
each individual lasso problem from the original dimension $d$ to the
sample size $n$, therefore a better efficiency can be achieved in
settings when $d \gg n$.  Based on our experiments, the speed up due
to the lossy screening rule can be up to $500\%$.  

Unlike the Meinshausen-B{\"{u}}hlmann graph approach, the graphical
lasso estimates the inverse covariance matrix. 
The lossless screening rule \citep{Witten11} greatly reduces the
computation required by the graphical lasso algorithm, especially when
the estimator is highly sparse.  The lossy screening rule can further
speed up the algorithm and provides an extra performance boost.

\section{Summary}

We developed a new package named \texttt{huge} for high dimensional
undirected graph estimation. The package is complementary to the existing \texttt{glasso} package by providing extra features and functional modules. We plan to maintain and support this package in the future.

\bibliography{HUGE}
\end{document}